\documentclass[11pt]{article}

\usepackage[final]{acl}
\usepackage{booktabs}
\usepackage{adjustbox}
\usepackage{times}
\usepackage{latexsym}

\usepackage[capitalise]{cleveref}
\crefname{figure}{Figure}{Figures}
\Crefname{figure}{Figure}{Figures}

\usepackage[T1]{fontenc}

\usepackage[utf8]{inputenc}

\usepackage{microtype}

\usepackage{inconsolata}

\usepackage{graphicx}
\usepackage{todonotes}
\title{AgentCompass: Towards Reliable Evaluation of Agentic Workflows in Production}

\author{\parbox{0.9\linewidth}{\centering{NVJK Kartik} \quad Garvit Sapra \quad Rishav Hada  \quad Nikhil Pareek \\
FutureAGI Inc. \\
\small{ \tt\{kartik.nvj,garvit.sapra,rishav,nikhil\}@futureagi.com}} }

\begin{document}
\maketitle
\begin{abstract}
With the growing adoption of Large Language Models (LLMs) in automating complex, multi-agent workflows, organizations face mounting risks from errors, emergent behaviors, and systemic failures that current evaluation methods fail to capture. We present AgentCompass, the first evaluation framework designed specifically for post-deployment monitoring and debugging of agentic workflows. AgentCompass models the reasoning process of expert debuggers through a structured, multi-stage analytical pipeline: error identification and categorization, thematic clustering, quantitative scoring, and strategic summarization. The framework is further enhanced with a dual memory system\textemdash episodic and semantic\textemdash that enables continual learning across executions. Through collaborations with design partners, we demonstrate the framework’s practical utility on real-world deployments, before establishing its efficacy against the publicly available TRAIL benchmark. AgentCompass achieves state-of-the-art results on key metrics, while uncovering critical issues missed in human annotations, underscoring its role as a robust, developer-centric tool for reliable monitoring and improvement of agentic systems in production.
\end{abstract}

\section{Introduction}

With recent advancements in Large Language Model's reasoning capabilities, they are increasingly being used for complex reasoning tasks such as agentic workflows. Agentic workflows allow users to automate certain tasks, based on internal thinking and reasoning capabilites of LLMs. These workflows range from being extremely simple such as retrieving a knowledge-base answer to a customer query, to very complex such as multi-step decision making in supply chain optimization or coordinating multiple specialized agents in a financial analysis pipeline.

Recent surveys by Wakefield Research \cite{pagerduty_agentic_ai_2025} and other organizations \cite{einpresswire_agentic_ai_market_2025} show, agentic workflows have surged dramatically across industries in 2025, with statistics showing widespread adoption, hefty return on investment (ROI) gains, and rapid scaling in enterprise environments. Organizations are deploying agentic workflows across critical business functions. Automation enabled by agentic workflows has resulted in 20–30\% cost savings for organizations, freeing resources for strategic growth \cite{belcak2025smalllanguagemodelsfuture, Yu_2025}. However, there are some major roadblocks in widespread adoption of agentic workflows. Organizations face a range of technical, organizational, and operational roadblocks. Agents often make decisions using non-standardized paths, destabilizing established workflows and making process oversight difficult in regulated environments. Organizations are concerned about bias, and poor-quality or fragmented datasets slow reliable adoption and amplify systemic errors. Poor evaluation and systems breaking in production are major causes of financial and reputational damage to organizations adopting agentic AI \cite{wang2025adaptiveaiagentplacement, liu2025realbarrierllmagent, chaudhry2025murakkabresourceefficientagenticworkflow}. Insufficient evaluation methods, over-reliance on technical benchmarks, and inadequate real-world robustness testing often leave organizations unprepared for edge cases and contextual failures. Most current evaluation frameworks prioritize technical metrics (accuracy, speed) and neglect stress tests for human-centered context, edge cases, and emotional intelligence, which are critical for user experience and trust. Errors and biases often compound and propagate through multi-agent workflows, making correction and accountability complex. Unaddressed failures can cause legal liabilities and loss of competitive advantage \cite{moteki2025fieldworkarenaagenticaibenchmark, meimandi2025measurementimbalanceagenticai, raza2025trismagenticaireview, yehudai2025surveyevaluationllmbasedagents, shukla2025adaptivemonitoringrealworldevaluation}.

To address this gap, recent research focuses on developing robust evaluation frameworks, that not only measure accuracy on a controlled dataset, but also includes real world scenarios with subjective evaluation metrics. For example, GAIA proposes a benchmark  for General AI Assistants \cite{mialon2023gaiabenchmarkgeneralai}.  GAIA focuses on real-world tasks that require reasoning, multi-modality, web browsing, and effective use of external tools. This moves beyond traditional benchmarks, which often focus on narrow tasks (like text classification or image recognition), by emphasizing an AI agent’s ability to generalize, adapt to changing scenarios, perform complex multi-step operations, and interact with diverse information sources. TRAIL \cite{deshpande2025trailtracereasoningagentic} extends the GAIA benchmark by providing an error taxonomy for agentic executions with an annotated dataset demonstrating the practical usage of the proposed taxonomy. 

Most organizations are unprepared for the complexities of agentic and multi-agentic AI risks, with governance blind spots multiplying post-deployment. These gaps often manifest as cascading failures, emergent behaviors, and unanticipated interactions with external systems, which existing monitoring and control frameworks struggle to detect or mitigate. In this work, we present AgentCompass, the first evaluation framework purpose-built for monitoring and debugging agentic workflows in production. Unlike prior approaches that rely on static benchmarks or single-pass LLM judgments, AgentCompass employs a recursive plan-and-execute reasoning cycle, a formal hierarchical error taxonomy tailored to business-critical issues, and a dual memory system that enables longitudinal analysis across executions. Its multi-stage analytical pipeline spans four key workflows: error identification and categorization, thematic error clustering, quantitative quality scoring, and synthesis with strategic summarization. To capture systemic risks, the framework further performs trace-level clustering via density-based analysis, surfacing recurring failure patterns that can be prioritized by engineering teams. Empirically, we validated AgentCompass with design partners on real-world deployments and demonstrated its generality through the publicly available TRAIL benchmark, where it achieves state-of-the-art results in error localization and joint metrics while uncovering critical errors that human annotations miss, including safety and reflection gaps. Together, these contributions make AgentCompass a rigorous and practical foundation for real-time production monitoring and continual improvement of complex agentic systems.



\section{Methodology}
\label{sec:methods}

We propose an agentic framework for the automated analysis of traces. Our approach models the cognitive workflow of an expert human debugger, creating a structured, multi-stage analytical pipeline for data that is unstructured as traces stored in open-telemetry format. The core of our method is a recursive reasoning framework that enhances the reliability of Large Language Models for complex diagnostic tasks. This framework is augmented with a knowledge persistence layer for continual learning. Our method includes 3 core components: a multi-stage analytical pipeline, trace-level issue clustering, and knowledge persistence for continual learning. We describe each of these components in detail in this section.

\subsection{The Multi-Stage Analytical Pipeline}
\label{ssec:agentic}
Our method deconstructs the complex goal of trace analysis into a sequence of four distinct, progressively abstract stages. Each stage's output serves as the structured input for the next, creating a coherent analytical narrative from granular findings to a high-level strategic summary.

\begin{enumerate}
    \item \textbf{Error Identification and Categorization:} The initial stage performs a comprehensive scan of the entire execution trace to identify discrete errors. Each identified error is classified according to a formal, hierarchical error taxonomy, grounding the analysis in a predefined set of failure modes common to agentic systems. This taxonomy is described in detail in section \ref{ssec:taxonomy}.

    \item \textbf{Thematic Error Clustering:} The discrete errors identified in the previous stage are then subjected to a thematic analysis. The system groups individual errors into semantically coherent clusters, designed to reveal systemic issues, causal chains, or recurring patterns of failure that would not be apparent from isolated error events.

    \item \textbf{Quantitative Quality Scoring:} To move beyond qualitative error descriptions, this stage assesses the overall quality of the trace across several predefined dimensions (e.g., factual grounding, safety, plan execution). The system assigns a quantitative score to each dimension, providing a multi-faceted, objective measure of the agent's performance.

    \item \textbf{Synthesis and Strategic Summarization:} The final stage synthesizes all preceding data \textemdash individual errors, thematic clusters, and quantitative scores into a final, actionable summary. This includes a single, aggregate quality score, key insights into the agent's behavior, and a recommended priority level for human intervention.
\end{enumerate}

\subsubsection{A Formal Taxonomy for Agentic Errors}
\label{ssec:taxonomy}
Underpinning our entire methodology is a formal, hierarchical error taxonomy that provides a structured ontology for classifying failures in agentic systems. This taxonomy serves as the analytical foundation for the Error Identification stage and as a guiding schema for the agent's reasoning. We take inspiration from the taxonomy proposed in TRAIL \cite{deshpande2025trailtracereasoningagentic}. 
Our taxonomy is designed to be comprehensive in its coverage, developer-centric in its focus, and includes key business critical issues, categorizing errors in a way that maps directly to actionable interventions.
\begin{figure*}[htbp]
  \centering
  \includegraphics[width=0.95\linewidth]{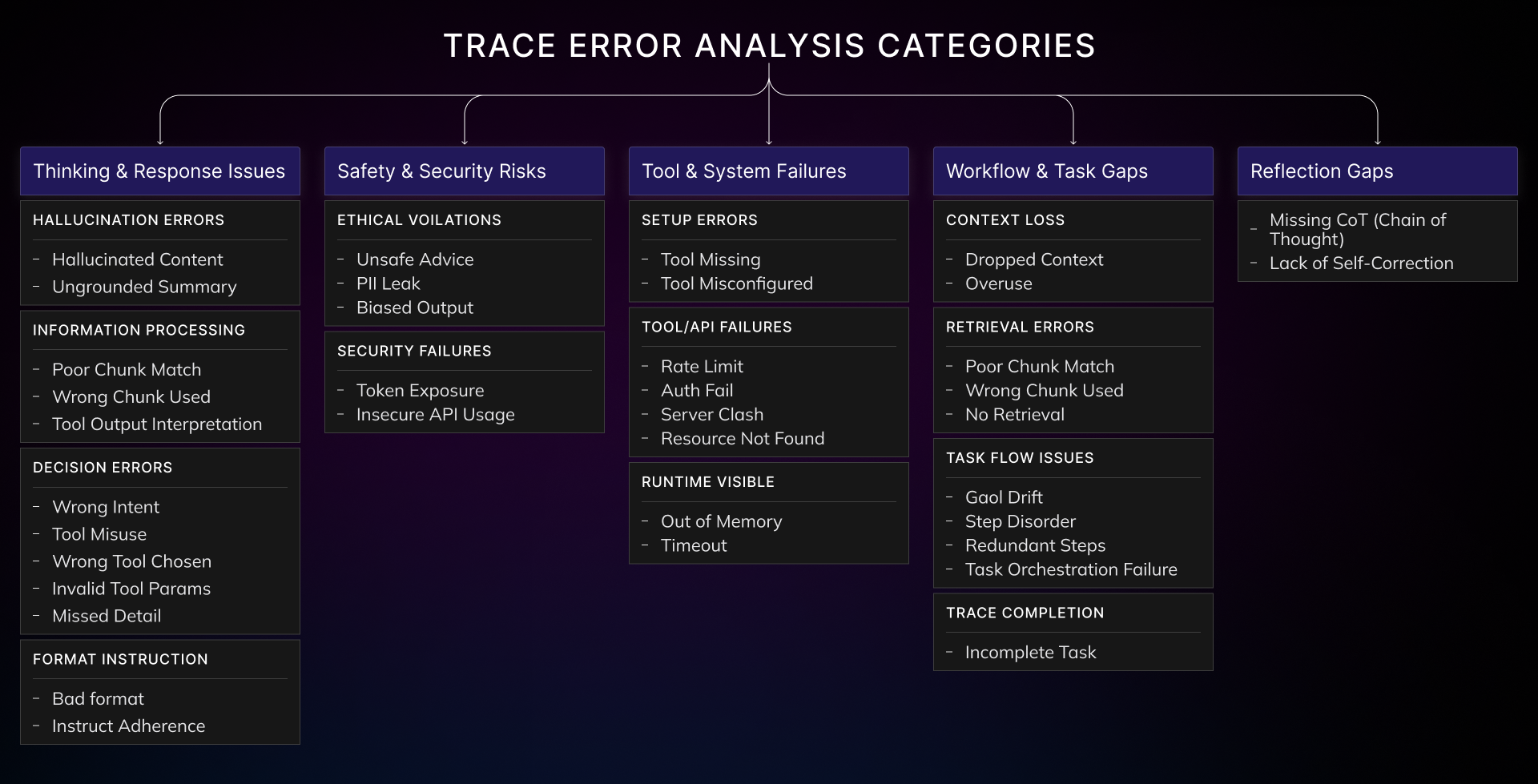}
  \caption{Detailed error taxonomy used in AgentCompass}
  \label{fig:taxonomy}
\end{figure*}
The taxonomy is organized into five high-level categories:
\begin{itemize}
    \item \textbf{Thinking \& Response Issues:} Failures in the agent's core reasoning, such as hallucinations, misinterpretation of retrieved information, flawed decision-making, or violations of output formatting constraints.
    \item \textbf{Safety \& Security Risks:} Behaviors that could lead to harm, including the leakage of personally identifiable information (PII), the exposure of credentials, or the generation of biased or unsafe content.
    \item \textbf{Tool \& System Failures:} Errors originating from the agent's interaction with external tools or its execution environment. This includes API failures, misconfigurations, rate limits, and runtime exceptions.
    \item \textbf{Workflow \& Task Gaps:} Breakdowns in the execution of multi-step tasks, such as the loss of conversational context, goal drift, inefficient or redundant actions, and failures in task orchestration.
    \item \textbf{Reflection Gaps:} A meta-level category for failures in the agent's introspective or planning capabilities, such as a lack of self-correction after an error or taking action without evidence of reasoning (e.g., Chain-of-Thought).
\end{itemize}
Each category is further broken down into specific subcategories and granular error types, providing a multi-level classification schema that enables both high-level thematic analysis and fine-grained root-cause identification. Detailed taxonomy is shown in figure \ref{fig:taxonomy}. 

\subsubsection{The Agentic Reasoning Framework}
For our agentic workflows described in section \ref{ssec:agentic} we adopt the ReAct framework \cite{yao2023reactsynergizingreasoningacting}. This framework is designed to steer and structure the inferential capabilities of LLMs.

\paragraph{The Plan-and-Execute Reasoning Cycle.}
At the heart of our methodology is a \textit{plan-and-execute} cognitive cycle. Rather than tasking an LLM with solving a complex problem in a single step as described in TRAIL using LLM-as-a-judge, we decompose the reasoning for each pipeline stage into two distinct phases. First, in the \textbf{Planning Phase}, the agent uses an LLM to generate a structured, explicit strategy for the task at hand (e.g., "identify critical spans to analyze for tool-use errors"). Second, in the \textbf{Execution Phase}, the agent provides this strategy back to the LLM as a directive for performing the actual analysis. This decomposition enforces a methodical approach, mitigates the risk of goal drift, and significantly improves the reliability and consistency of the analytical output compared to single-pass prompting.

\paragraph{Structured Representation of Trace Data.}
To make complex, non-linear execution traces legible to a language model, our method first transforms the raw trace data. It reconstructs the causal hierarchy of execution spans into a tree structure, which is then serialized into a numbered, hierarchical text format. This representation provides the LLM with both a high-level "outline" of the execution and detailed, step-by-step "details" for deeper inspection, effectively creating a structured document for the model to analyze.

\subsection{Trace-level Issue Clustering via Density-Based Analysis}
To provide developers with a higher-level understanding of recurring issues across multiple traces and over time, our methodology includes a cross-trace clustering stage. This moves beyond single-trace analysis to identify repeated failure modes that can help the developers identify and prevent it in future. Instead of simple rule-based grouping, we employ a sophisticated, unsupervised machine learning approach.

First, each identified error is converted into a high-dimensional vector representation using a transformer based encoder model. These embeddings are generated from a concatenation of some of the error's key semantic features.

Next, we apply the Hierarchical Density-Based Spatial Clustering of Applications with Noise (HDBSCAN) algorithm to this collection of error embeddings. We chose HDBSCAN for its ability to discover clusters of varying shapes and densities and its robustness to noise, making it well-suited for the diverse and often sparse nature of error data. The algorithm groups semantically similar error embeddings into dense clusters, effectively identifying groups of recurring, related issues.

Finally, to account for ambiguous cases, we employ a soft clustering technique. Errors that are initially classified as noise by HDBSCAN are probabilistically assigned to the most suitable existing cluster if their membership probability exceeds a predefined threshold. This allows the system to group peripherally related errors without forcing them into ill-fitting categories or just forming a singleton cluster, providing a more nuanced and comprehensive view of systemic issues. The resulting clusters represent high-level "issues" that can be tracked and prioritized by engineering teams in the form of `tickets' or `issues' which can be tracked in their agile development cycle.

\subsection{Knowledge Persistence for Continual Learning}
Most evaluation frameworks treat each agentic execution as independent, overlooking recurring error patterns and domain-specific nuances. In production, however, deployed agents repeatedly encounter similar edge cases, systemic errors, and domain-specific critical issues that cannot be fully anticipated at design time. By equipping our framework with memory, we enable longitudinal analysis: allowing the system to contextualize new traces with prior findings and continually refine its diagnostic heuristics, thereby improving accuracy and business relevance over time.

\paragraph{Episodic and Semantic Memory.}
The system maintains two distinct memory stores. \textbf{Episodic Memory} captures context and findings related to specific, individual traces, allowing for stateful, multi-turn analysis of a single execution. \textbf{Semantic Memory}, in contrast, stores generalized, cross-trace knowledge. The system abstracts high-confidence findings into durable error patterns or best practices, allowing it to recognize recurring issues across a fleet of agents and improve its diagnostic acumen over time.

\paragraph{Memory-Augmented Reasoning.}
Before initiating analysis on a new trace, the agent queries its memory stores for relevant prior knowledge. This retrieved context is then injected directly into the prompts used during the plan-and-execute cycle, priming the LLM with historical context and learned heuristics to guide its analysis.

\section{Evaluation Study}
\label{sec:evaluation}

To empirically validate the effectiveness of our agentic framework, we first collaborated with several customers as design partners, testing the system on real-world scenarios and validating its outputs against proprietary datasets. This early deployment phase demonstrated strong accuracy and practical utility in production-like settings. To further establish the general efficacy of our approach, we then evaluated it against the publicly available, human-annotated TRAIL benchmark. Our objectives were twofold: (i) to quantify performance in error identification and categorization against a standardized ground truth, and (ii) to qualitatively assess the framework’s ability to surface valid errors overlooked during manual annotation.

\subsection{Benchmark Dataset}
For our evaluation, we utilized the TRAIL (Trace Reasoning and Agentic Issue Localization) benchmark \cite{deshpande2025trailtracereasoningagentic}, a publicly available dataset specifically designed for evaluating trace analysis systems. The dataset consists of 148 traces totaling 1,987 OpenTelemetry spans, of which 575 exhibit at least one error, sourced from two established agent benchmarks: GAIA \cite{mialon2023gaiabenchmarkgeneralai}, which focuses on open-world information retrieval, and SWE-Bench \cite{jimenez2024swebenchlanguagemodelsresolve}, which involves software engineering tasks.

Each trace in the TRAIL dataset is accompanied by a set of human-generated annotations. These annotations identify the precise location (span ID) and the specific category of each error according to the taxonomy proposed in their work.

\subsection{Experimental Setup}
Our experimental setup was designed to measure the agent's performance along the two primary axes defined by the TRAIL benchmark: error categorization and localization.

We ran our framework, with its full plan-and-execute and memory-augmented reasoning capabilities using our proprietary in-house fine-tuned Turing Large model\footnote{\url{https://docs.futureagi.com/future-agi/get-started/evaluation/future-agi-models}}, on each of the 148 traces from the TRAIL dataset. The agent's output was again then compared with human annotated ground truth.

\paragraph{Taxonomy Alignment.}
As our agent employs a more granular, hierarchical taxonomy (detailed in Section \ref{ssec:taxonomy}), we first established a semantic mapping to the TRAIL taxonomy to ensure a fair comparison. For example, multiple specific errors from our agent's taxonomy, such as \textit{`Hallucinated Content'} or \textit{`Hallucinated Tool Result'} were mapped to the corresponding TRAIL category of \textit{`Language-only'}. This normalization enabled a direct and meaningful comparison of categorization accuracy.

From these counts, we calculated the overall Categorization F1-Score, providing a single, balanced measure of our agent's performance that is directly comparable to the results reported for other models in the TRAIL study. We also specifically isolated the set of False Positives for a subsequent qualitative analysis to identify valid errors our agent found that were not in the original benchmark.

\section{Results and Analysis}
\label{sec:results}

\begin{table*}[!ht]
    \centering
    \small
    \setlength{\tabcolsep}{4pt}
    \begin{tabular}{l c c c c c c c c}
        \toprule
         & \multicolumn{4}{c}{\textsc{TRAIL} (GAIA)} & \multicolumn{4}{c}{\textsc{TRAIL} (SWE Bench)} \\
         \cmidrule(lr){2-5} \cmidrule(lr){6-9}
         Model & Cat. F1 & Loc. Acc. & Joint & $\rho$ & Cat. F1 & Loc. Acc. & Joint & $\rho$ \\\midrule
         \textsc{Llama-4-Scout-17B-16E-Instruct} & 0.041 & 0.000 & 0.000 & 0.134 & 0.050 & 0.000 & 0.000 & 0.264 \\
         \textsc{Llama-4-Maverick-17B-128E-Instruct} &  0.122 & 0.023 & 0.000 & 0.338 & 0.191 & 0.083 & 0.000 & -0.273  \\
         \textsc{GPT-4.1} & 0.218 & 0.107 & 0.028 & 0.411 & 0.166 & 0.000 & 0.000 & 0.153\\
         \textsc{Open AI o1} & 0.138 & 0.040 & 0.013 & 0.450 & \texttt{CLE} & \texttt{CLE} & \texttt{CLE} & \texttt{CLE} \\
         \textsc{Open AI o3} & 0.296 & 0.535 & 0.092 & 0.449 & \texttt{CLE} & \texttt{CLE} & \texttt{CLE} & \texttt{CLE} \\
         \textsc{Anthropic Claude-3.7-Sonnet} & 0.254 & 0.204 & 0.047 & \textbf{0.738} & \texttt{CLE} & \texttt{CLE} & \texttt{CLE} & \texttt{CLE} \\
         \textsc{Gemini-2.5-Pro-Preview-05-06} & \textbf{0.389} & 0.546 & 0.183 & 0.462 & 0.148 & 0.238 & 0.050 & \textbf{0.817} \\
         \textsc{Gemini-2.5-Flash-Preview-04-17} & 0.337 & 0.372 & 0.100 & 0.550 & 0.213 & 0.060 & 0.000 & 0.292\\
         \textsc{FAGI-AgentCompass} & 0.309 & \textbf{0.657} & \textbf{0.239} & 0.430 & \textbf{0.232} & \textbf{0.250} & \textbf{0.051} & 0.408\\
         \bottomrule
    \end{tabular}
    \vspace{-0.2em}
    \caption{Performance across LLMs for Error Categorization \& Localization on TRAIL Benchmark. Note: Performance values for models other than FAGI-AgentCompass have been taken from \citet{deshpande2025trailtracereasoningagentic}.}
    \vspace{-0.2em}
    \label{tab:trail_performance}
\end{table*}

Our evaluation demonstrates that the \texttt{FAGI-AgentCompass} framework achieves state-of-the-art performance on the TRAIL benchmark, significantly outperforming existing general-purpose Large Language Models on key metrics for trace debugging. The complete performance comparison is presented in Table~\ref{tab:trail_performance}.

\subsection{Quantitative Performance}
We report the performance on the same metrics that were used to evaluate the existing language models for the TRAIL Benchmark, Categorization F1-Score (Cat. F1), Localization Accuracy (Loc. Acc.), and the Joint score, which represents the accuracy of predicting both category and location correctly.

\paragraph{Localization Accuracy.}
The most significant result is our agent's superior performance in Localization Accuracy. On the TRAIL (GAIA split) dataset, \texttt{FAGI-AgentCompass} achieves a Localization Accuracy of \textbf{0.657}, substantially outperforming the next best model, Gemini-2.5-Pro, which scored 0.546. This indicates that our agent's iterative planning and executing cycle along with memory augmentation allows it to pinpoint the precise event or set of spans when an error occured. Similarly, on the more complex TRAIL (SWE Bench Split) dataset, our agent achieves the highest Localization Accuracy of \textbf{0.250}.

\paragraph{Categorization F1-Score.}
In terms of Categorization F1-Score, our agent is highly competitive, achieving 0.309 on GAIA. While the Gemini-2.5-Pro model reports a higher F1 score on this specific dataset, this metric is sensitive to the inherent differences between our agent's granular, hierarchical taxonomy and the one used for the TRAIL benchmark annotations. Our primary goal was to validate our agent's performance against the established benchmark, which necessitated mapping our more detailed categories to their broader counterparts. This translation can naturally introduce discrepancies that affect the metric. Despite this, our agent was able to achieve a well balanced and top-tier performance.
Furthermore, on the highly technical SWE Bench dataset, our agent again leads with the highest Categorization F1-Score of \textbf{0.232}, highlighting its robustness on code-centric traces where general models often struggle.

\paragraph{Joint Performance.}
The Joint score, which is the strictest metric, further underscores the efficacy of our approach. Our agent sets a new state-of-the-art on both datasets, with a score of \textbf{0.239} on GAIA and \textbf{0.051} on SWE Bench. This demonstrates that the agent's ability to localize errors is not at the expense of categorization accuracy. The plan-and-execute cycle, which first identifies critical spans and then analyzes them for specific error types, leads to a higher probability of getting both the location and category correct simultaneously.

\paragraph{Score Correlation.}
The Pearson correlation coefficient ($\rho$), which measures the correlation between our agent's and human's overall quality scores for a trace, provides further insight into its analytical behavior. \texttt{FAGI-AgentCompass} reports $\rho$ scores of 0.430 and 0.408 on GAIA and SWE Bench datasets respectively. Our framework employs a more systematic error taxonomy and broader evaluation criteria than those defined in the benchmark, which naturally leads to moderate (rather than high) correlation with human scores while capturing a richer spectrum of errors.

\begin{figure}[ht!]
  \centering
  \includegraphics[width=0.9\linewidth]{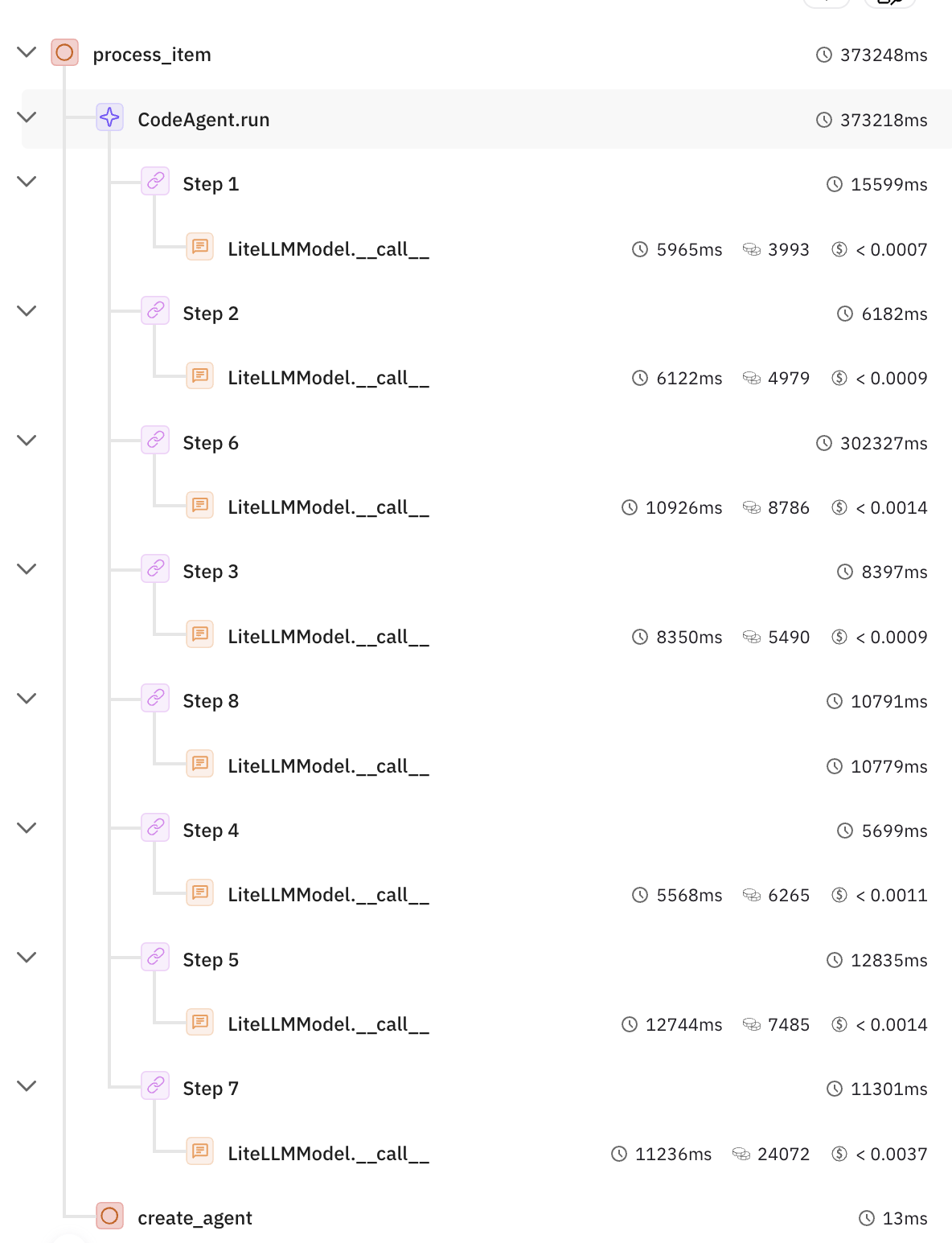}
  \caption{Execution trace for scenario 4}
  \label{fig:trace5}
\end{figure}

Additionally, a deeper analysis reveals that our agent identified errors that were missed by human judgment. For example as shown in figure \ref{fig:trace5} in table \ref{tab:qualitative_comparison} one of the categories in our taxonomy that the original benchmark study completely misses is \textit{`Safety and Security Alerts'} responsible for identifying outputs that may potentially cause harm, leak personal data or violate best security practices.
When our agent detects additional, valid errors, it correctly assigns a lower overall quality score to the trace. This necessary and more accurate assessment naturally diverges from the human score, which was based on an incomplete view of the trace's failures. 
Therefore, a moderate correlation is an expected and even desirable artifact of a system designed to be more rigorous and developer-centric than a manual annotation process using a limited taxonomy. Our agent's scoring is a function of the comprehensive set of issues it uncovers, not just a replication of a human's holistic impression.

\begin{table*}[!ht]
    \centering
    \small
    \begin{tabular}{p{1.5cm} p{6cm} p{6cm}}
        \toprule
        \textbf{Trace} & \textbf{Ground Truth Analysis Summary (from TRAIL)} & \textbf{FAGI-AgentCompass Analysis Summary} \\
        \midrule
        \texttt{\cref{fig:trace1}} & Identifies a \textbf{Tool Selection Error}, noting the agent failed to call the \texttt{search\_agent} and instead printed a task string. & Correctly identifies the same tool misuse but also uncovers a \textbf{Hallucination} where the agent invents a final answer, and a \textbf{Lack of Self-Correction} as it repeatedly fails to use the tool correctly before fabricating the result. \\
        \addlinespace
        \texttt{\cref{fig:trace2}} & Flags multiple \textbf{Formatting Errors} where the agent repeatedly calls the \texttt{page\_down} tool with incorrect parameters. & Identifies the same formatting errors but provides a much deeper diagnosis. It flags a critical \textbf{Lack of Self-Correction}, as the agent ignores explicit error messages for 10 consecutive steps, and notes that this violates existing knowledge in its \textbf{Semantic Memory}, indicating a systemic failure to learn. \\
        \addlinespace
        \texttt{\cref{fig:trace3}} & Notes a \textbf{Poor Information Retrieval} error, as the agent's web search returned an irrelevant result. & Pinpoints the sub-agent's tool failures (`Invalid Tool Params`, `Lack of Self-Correction`) that caused the retrieval to fail. More critically, it uncovers the parent agent's subsequent \textbf{Goal Drift} and severe \textbf{Hallucination}, where it fabricates a numerical answer after receiving the sub-agent's failure report. \\
        \addlinespace
        \texttt{\cref{fig:trace4}} & Identifies a \textbf{Tool Selection Error}, as the agent hallucinates a Tropicos ID instead of searching for it. & Correctly identifies the hallucination but also diagnoses a \textbf{Task Orchestration Failure} and a \textbf{Missing ReAct Planning} error. Our agent astutely notes the root cause: a failure to execute its own valid plan, which is a critical breakdown between the agent's reasoning and its subsequent actions. \\
        \addlinespace
        \texttt{\cref{fig:trace5}} & Identifies an \textbf{Instruction Non-compliance} error because the agent prints 1000 characters of a file when the limit was 500. & Our agent identifies the same violation but classifies it with much higher severity as a \textbf{Safety \& Security Risks > Data Exposure} error. It correctly reasons that violating an explicit negative constraint about data handling is a security failure, not just a formatting mistake, providing a more critical and accurate diagnosis. \\
        \bottomrule
    \end{tabular}
    \caption{Qualitative comparison of error analysis on select traces from the TRAIL benchmark. The FAGI-AgentCompass consistently provides a deeper, more actionable root-cause analysis by identifying planning and meta-level reasoning failures that supplement the ground truth annotations.}
    \label{tab:qualitative_comparison}
\end{table*}

\begin{figure}[ht!]
  \centering
  \includegraphics[width=0.9\linewidth]{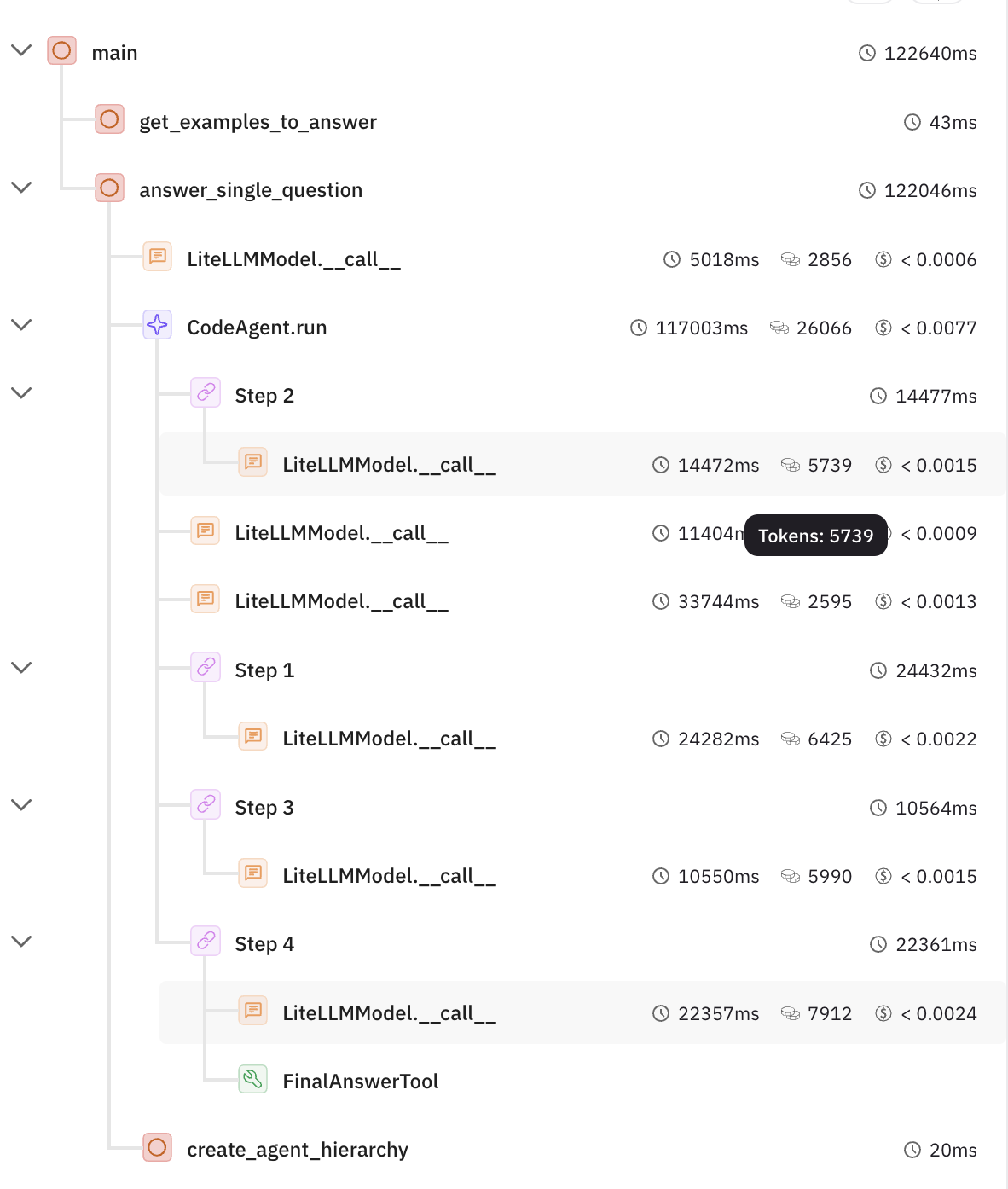}
  \caption{Execution trace for scenario 1}
  \label{fig:trace1}
\end{figure}

\begin{figure}[ht!]
  \centering
  \includegraphics[width=0.9\linewidth]{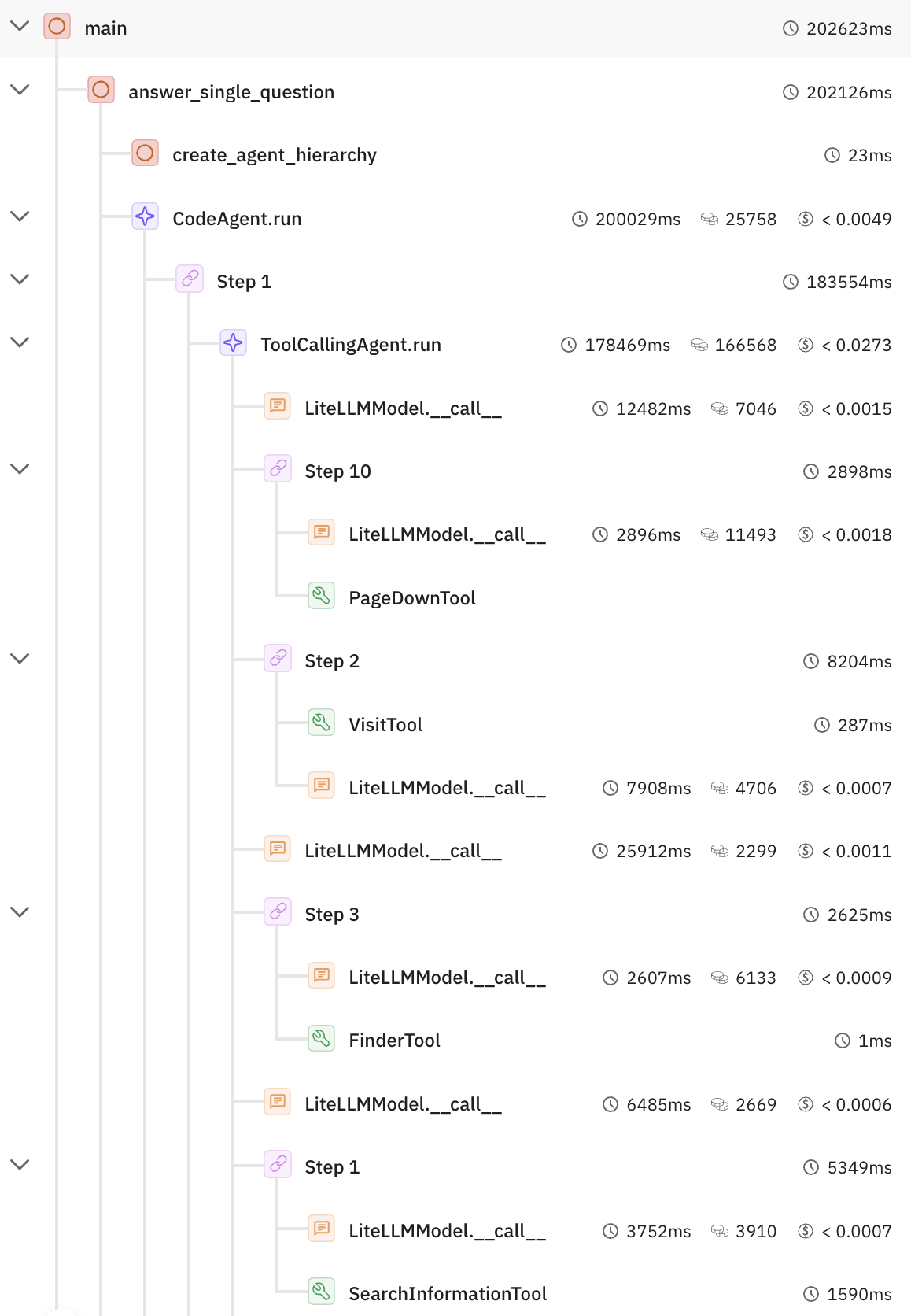}
  \caption{Execution trace for scenario 2}
  \label{fig:trace2}
\end{figure}

\begin{figure}[ht!]
  \centering
  \includegraphics[width=0.9\linewidth]{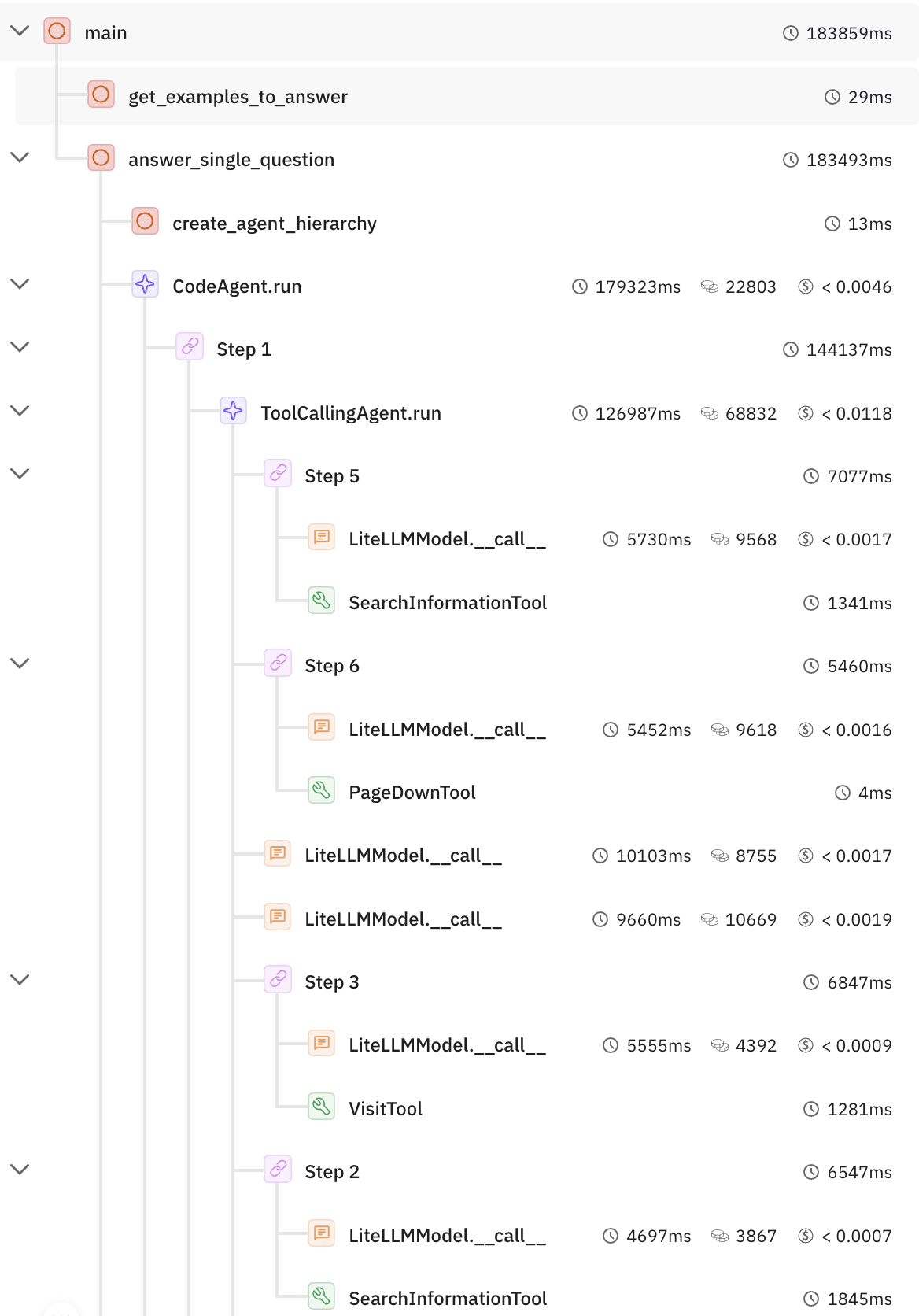}
  \caption{Execution trace for scenario 3}
  \label{fig:trace3}
\end{figure}

\begin{figure}[ht!]
  \centering
  \includegraphics[width=0.9\linewidth]{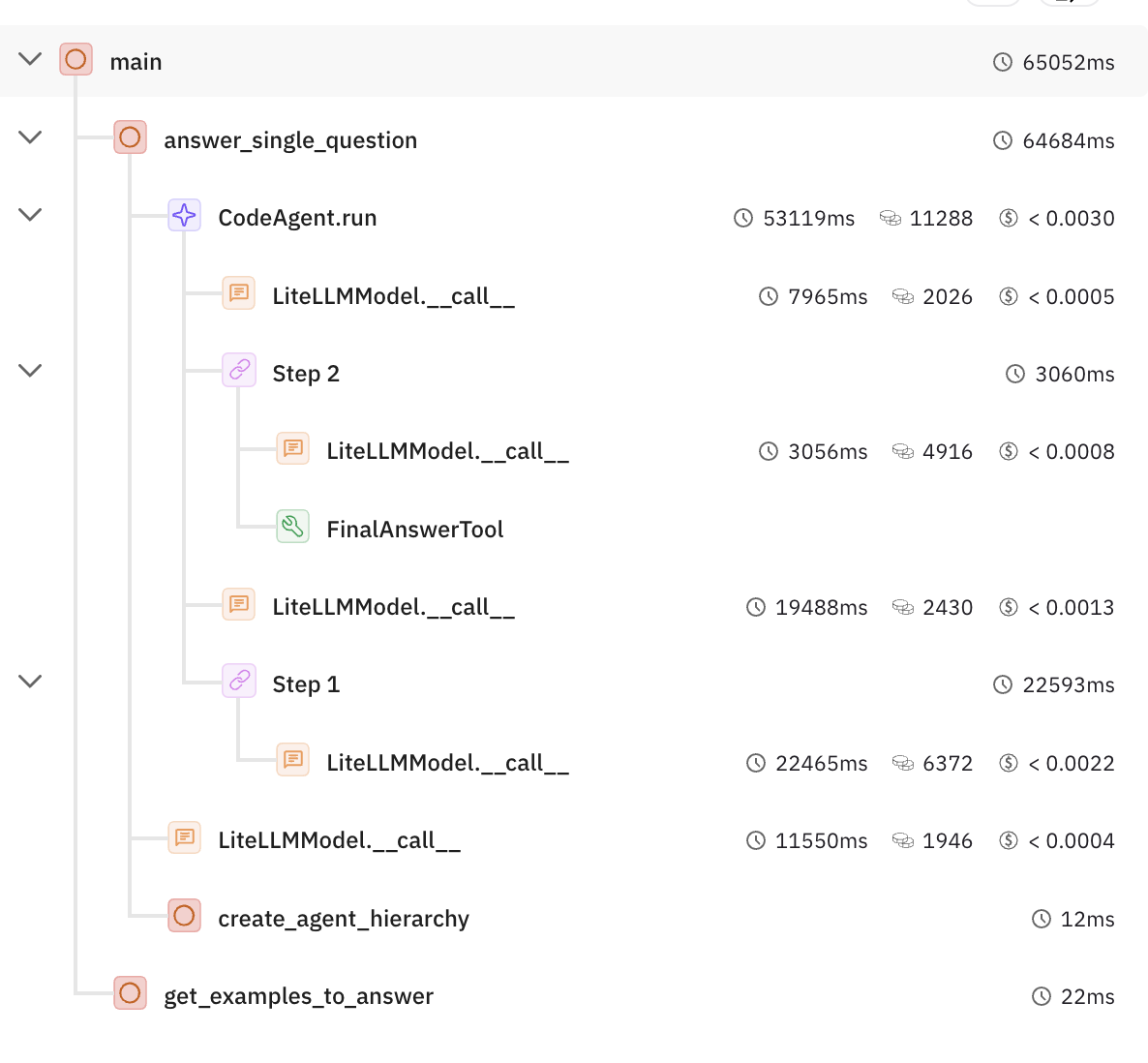}
  \caption{Execution trace for scenario 4}
  \label{fig:trace4}
\end{figure}

\begin{figure*}[ht!]
  \centering
  \includegraphics[width=1\linewidth]{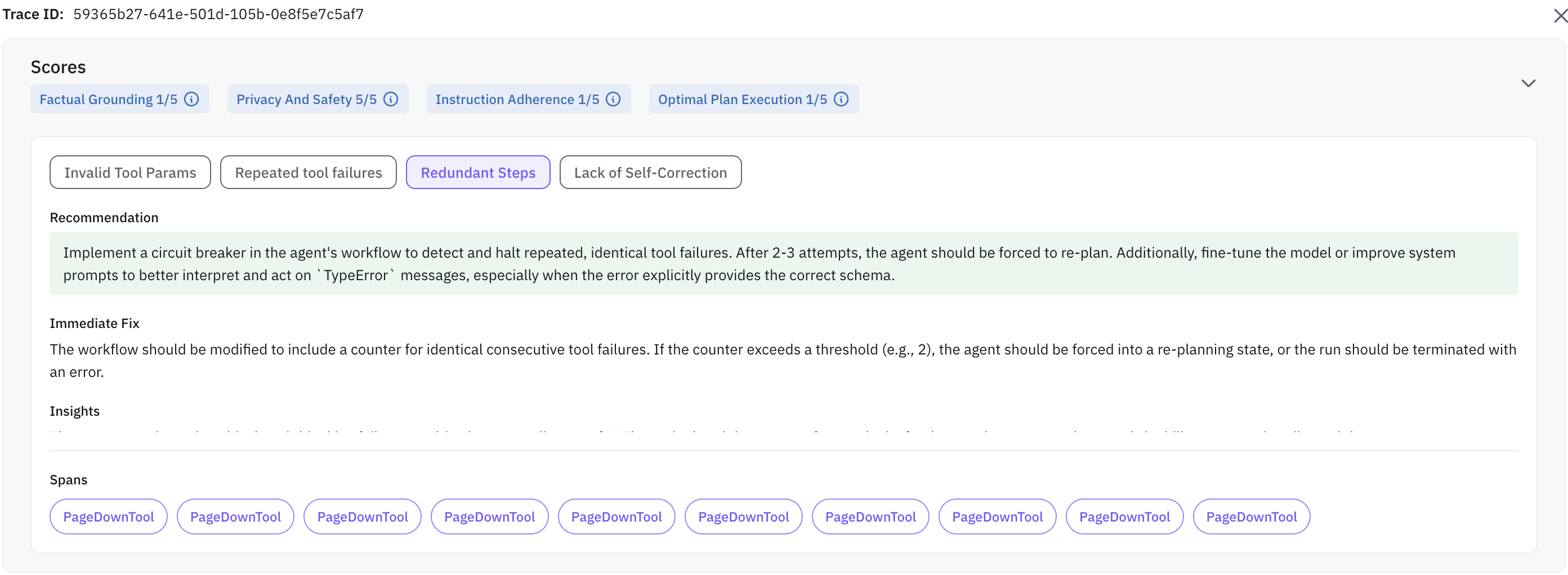}
  \caption{Recommended fixes by AgentCompass, example 1}
  \label{fig:Reccom1}
\end{figure*}
\begin{figure*}[ht!]
  \centering
  \includegraphics[width=1\linewidth]{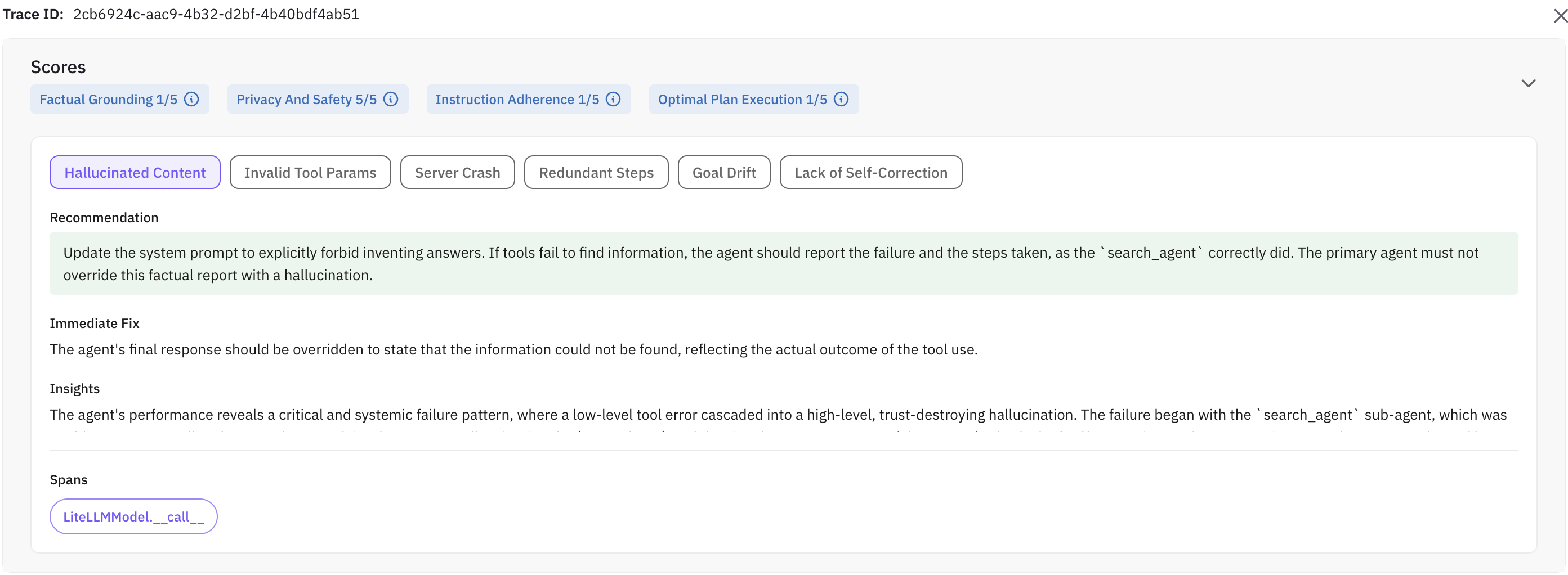}
  \caption{Recommended fixes by AgentCompass, example 2}
  \label{fig:Reccom2}
\end{figure*}

\subsection{Qualitative Analysis: Uncovering Novel Errors}
Beyond quantitative metrics, a key finding of our study is the agent's ability to identify valid errors that were not included in the original human-annotated ground truth. These findings correspond to the False Positives (FP) in our evaluation metric, and a manual review revealed that a significant portion of these were not erroneous agent predictions, but rather novel, valid insights.

We show a few samples in table \ref{tab:qualitative_comparison}. For instance, in one of the traces belonging to the benchmark dataset, the agent identified a \textit{`Tool Selection Error'} because the plan failed to specify which tool should be used to access external information from the corresponding database. While the human annotators did not label this as a distinct error, it represents a critical failure in agentic planning that leads to downstream hallucinations. Our agent's structured analysis of the entire reasoning chain, not just the final tool call, allows it to detect these subtle but crucial planning-phase failures that will potentially benefit the AI developers who might miss the underlying cause that triggers the issue in their agentic workflows.

As established in our analysis of the score correlation, our agent's comprehensive taxonomy allows it to identify critical issues missed during manual annotation, such as the \textit{`Safety \& Security Risks`} previously discussed.

In addition to security, our agent demonstrated a strong capability for identifying subtle failures in the agent's cognitive processes, a crucial area for modern reasoning-based systems. Our taxonomy includes a dedicated category for \textit{`Reflection Gaps'}, which covers failures in an agent's introspective and planning capabilities. For example, in several complex traces, our agent flagged a \textit{`Lack of Self-Correction'}. It precisely identified moments where a sub-agent, after receiving an error from a tool, would immediately retry the exact same failed action without modifying its parameters or approach. This type of recursive, inefficient behavior is a significant flaw in agentic logic that a less granular analysis might miss. By identifying not just the direct tool failure but the meta-level reasoning failure that caused it, our framework provides a more profound and actionable diagnosis for developers seeking to build more robust and intelligent agents. Our framework not only identifies these errors, but also suggests actionable fixes to improve the system. Figures \ref{fig:Reccom1} and \ref{fig:Reccom2} show representative examples where the agent produces prescriptive “Fix Recipes” that map detected errors to targeted remediation strategies, such as adjusting tool invocation parameters, refining retrieval prompts, or restructuring orchestration logic. These recommendations provide developers with concrete next steps, reducing debugging effort and accelerating the deployment of more reliable agentic workflows.

\section{Conclusion}
In this work, we introduced AgentCompass, a memory-augmented evaluation framework for diagnosing and improving agentic workflows post-deployment. By combining a structured error taxonomy, multi-stage analytical reasoning, trace-level clustering, and persistent memory, our system provides deeper and more actionable insights than existing evaluation approaches. Our empirical study demonstrates strong accuracy in real-world deployments and state-of-the-art performance on the TRAIL benchmark, including the ability to uncover valid errors overlooked by human annotation. These findings highlight that moderate correlation with human judgments reflects not a shortcoming but a more rigorous, systematic evaluation process that captures the full spectrum of agentic failures. As organizations scale their use of multi-agent AI systems, AgentCompass offers a path toward trustworthy, production-grade monitoring and debugging, bridging the gap between technical benchmarks and the realities of enterprise deployment.

\newpage
\bibliography{acl_latex}

\begin{thebibliography}{16}
\providecommand{\natexlab}[1]{#1}

\bibitem[{Belcak et~al.(2025)Belcak, Heinrich, Diao, Fu, Dong, Muralidharan, Lin, and Molchanov}]{belcak2025smalllanguagemodelsfuture}
Peter Belcak, Greg Heinrich, Shizhe Diao, Yonggan Fu, Xin Dong, Saurav Muralidharan, Yingyan~Celine Lin, and Pavlo Molchanov. 2025.
\newblock \href {https://arxiv.org/abs/2506.02153} {Small language models are the future of agentic ai}.
\newblock \emph{Preprint}, arXiv:2506.02153.

\bibitem[{Chaudhry et~al.(2025)Chaudhry, Choukse, Qiu, Íñigo Goiri, Fonseca, Belay, and Bianchini}]{chaudhry2025murakkabresourceefficientagenticworkflow}
Gohar~Irfan Chaudhry, Esha Choukse, Haoran Qiu, Íñigo Goiri, Rodrigo Fonseca, Adam Belay, and Ricardo Bianchini. 2025.
\newblock \href {https://arxiv.org/abs/2508.18298} {Murakkab: Resource-efficient agentic workflow orchestration in cloud platforms}.
\newblock \emph{Preprint}, arXiv:2508.18298.

\bibitem[{Deshpande et~al.(2025)Deshpande, Gangal, Mehta, Krishnan, Kannappan, and Qian}]{deshpande2025trailtracereasoningagentic}
Darshan Deshpande, Varun Gangal, Hersh Mehta, Jitin Krishnan, Anand Kannappan, and Rebecca Qian. 2025.
\newblock \href {https://arxiv.org/abs/2505.08638} {Trail: Trace reasoning and agentic issue localization}.
\newblock \emph{Preprint}, arXiv:2505.08638.

\bibitem[{{EIN Presswire / Market.us}(2025)}]{einpresswire_agentic_ai_market_2025}
{EIN Presswire / Market.us}. 2025.
\newblock Agentic ai market boosts by performing self actions grows by usd 196.6 billion by 2034, region at usd 1.58 billion.
\newblock \url{https://www.einpresswire.com/article/782914390/}.
\newblock Press release, February 4, 2025. Accessed: 2025-09-17.

\bibitem[{Jimenez et~al.(2024)Jimenez, Yang, Wettig, Yao, Pei, Press, and Narasimhan}]{jimenez2024swebenchlanguagemodelsresolve}
Carlos~E. Jimenez, John Yang, Alexander Wettig, Shunyu Yao, Kexin Pei, Ofir Press, and Karthik Narasimhan. 2024.
\newblock \href {https://arxiv.org/abs/2310.06770} {Swe-bench: Can language models resolve real-world github issues?}
\newblock \emph{Preprint}, arXiv:2310.06770.

\bibitem[{Liu et~al.(2025)Liu, Qin, Huang, Zeng, Xi, Lin, Wu, Wang, Shang, Tang, Lian, Yu, and Zhang}]{liu2025realbarrierllmagent}
Weiwen Liu, Jiarui Qin, Xu~Huang, Xingshan Zeng, Yunjia Xi, Jianghao Lin, Chuhan Wu, Yasheng Wang, Lifeng Shang, Ruiming Tang, Defu Lian, Yong Yu, and Weinan Zhang. 2025.
\newblock \href {https://arxiv.org/abs/2505.17767} {The real barrier to llm agent usability is agentic roi}.
\newblock \emph{Preprint}, arXiv:2505.17767.

\bibitem[{Meimandi et~al.(2025)Meimandi, Aránguiz-Dias, Kim, Saadeddin, and Kochenderfer}]{meimandi2025measurementimbalanceagenticai}
Kiana~Jafari Meimandi, Gabriela Aránguiz-Dias, Grace~Ra Kim, Lana Saadeddin, and Mykel~J. Kochenderfer. 2025.
\newblock \href {https://arxiv.org/abs/2506.02064} {The measurement imbalance in agentic ai evaluation undermines industry productivity claims}.
\newblock \emph{Preprint}, arXiv:2506.02064.

\bibitem[{Mialon et~al.(2023)Mialon, Fourrier, Swift, Wolf, LeCun, and Scialom}]{mialon2023gaiabenchmarkgeneralai}
Grégoire Mialon, Clémentine Fourrier, Craig Swift, Thomas Wolf, Yann LeCun, and Thomas Scialom. 2023.
\newblock \href {https://arxiv.org/abs/2311.12983} {Gaia: a benchmark for general ai assistants}.
\newblock \emph{Preprint}, arXiv:2311.12983.

\bibitem[{Moteki et~al.(2025)Moteki, Masui, Yang, Song, Bisk, Neubig, Kusajima, Watanabe, Ishida, Takahashi, and Jiang}]{moteki2025fieldworkarenaagenticaibenchmark}
Atsunori Moteki, Shoichi Masui, Fan Yang, Yueqi Song, Yonatan Bisk, Graham Neubig, Ikuo Kusajima, Yasuto Watanabe, Hiroyuki Ishida, Jun Takahashi, and Shan Jiang. 2025.
\newblock \href {https://arxiv.org/abs/2505.19662} {Fieldworkarena: Agentic ai benchmark for real field work tasks}.
\newblock \emph{Preprint}, arXiv:2505.19662.

\bibitem[{{PagerDuty}(2025)}]{pagerduty_agentic_ai_2025}
{PagerDuty}. 2025.
\newblock 2025 agentic ai roi survey results.
\newblock \url{https://www.pagerduty.com/resources/ai/learn/companies-expecting-agentic-ai-roi-2025/}.
\newblock Accessed: 2025-09-17.

\bibitem[{Raza et~al.(2025)Raza, Sapkota, Karkee, and Emmanouilidis}]{raza2025trismagenticaireview}
Shaina Raza, Ranjan Sapkota, Manoj Karkee, and Christos Emmanouilidis. 2025.
\newblock \href {https://arxiv.org/abs/2506.04133} {Trism for agentic ai: A review of trust, risk, and security management in llm-based agentic multi-agent systems}.
\newblock \emph{Preprint}, arXiv:2506.04133.

\bibitem[{Shukla(2025)}]{shukla2025adaptivemonitoringrealworldevaluation}
Manish Shukla. 2025.
\newblock \href {https://arxiv.org/abs/2509.00115} {Adaptive monitoring and real-world evaluation of agentic ai systems}.
\newblock \emph{Preprint}, arXiv:2509.00115.

\bibitem[{Wang et~al.(2025)Wang, He, Tang, Guo, Lou, Qian, Wang, and Jia}]{wang2025adaptiveaiagentplacement}
Xingdan Wang, Jiayi He, Zhiqing Tang, Jianxiong Guo, Jiong Lou, Liping Qian, Tian Wang, and Weijia Jia. 2025.
\newblock \href {https://arxiv.org/abs/2508.03345} {Adaptive ai agent placement and migration in edge intelligence systems}.
\newblock \emph{Preprint}, arXiv:2508.03345.

\bibitem[{Yao et~al.(2023)Yao, Zhao, Yu, Du, Shafran, Narasimhan, and Cao}]{yao2023reactsynergizingreasoningacting}
Shunyu Yao, Jeffrey Zhao, Dian Yu, Nan Du, Izhak Shafran, Karthik Narasimhan, and Yuan Cao. 2023.
\newblock \href {https://arxiv.org/abs/2210.03629} {React: Synergizing reasoning and acting in language models}.
\newblock \emph{Preprint}, arXiv:2210.03629.

\bibitem[{Yehudai et~al.(2025)Yehudai, Eden, Li, Uziel, Zhao, Bar-Haim, Cohan, and Shmueli-Scheuer}]{yehudai2025surveyevaluationllmbasedagents}
Asaf Yehudai, Lilach Eden, Alan Li, Guy Uziel, Yilun Zhao, Roy Bar-Haim, Arman Cohan, and Michal Shmueli-Scheuer. 2025.
\newblock \href {https://arxiv.org/abs/2503.16416} {Survey on evaluation of llm-based agents}.
\newblock \emph{Preprint}, arXiv:2503.16416.

\bibitem[{Yu et~al.(2025)Yu, Cheng, Cui, Gao, Luo, Wang, Zheng, and Zhao}]{Yu_2025}
Chaojia Yu, Zihan Cheng, Hanwen Cui, Yishuo Gao, Zexu Luo, Yijin Wang, Hangbin Zheng, and Yong Zhao. 2025.
\newblock \href {https://doi.org/10.1109/icaibd64986.2025.11082076} {A survey on agent workflow – status and future}.
\newblock In \emph{2025 8th International Conference on Artificial Intelligence and Big Data (ICAIBD)}, page 770–781. IEEE.

\end{thebibliography}




\end{document}